\documentclass{article}

\usepackage[preprint]{neurips_2025}

\usepackage[utf8]{inputenc} 
\usepackage[T1]{fontenc}    
\usepackage{hyperref}       
\usepackage{url}            
\usepackage{booktabs}       
\usepackage{amsfonts}       
\usepackage{nicefrac}       
\usepackage{microtype}      
\usepackage{xcolor}         
\usepackage{graphicx}
\usepackage{url}
\usepackage{hyperref}
\usepackage{booktabs}
\usepackage{pifont}
\newcommand{\cmark}{\ding{51}}  
\newcommand{\xmark}{\ding{55}}  
\usepackage{graphicx}
\usepackage[table]{xcolor}  
\usepackage{colortbl}       
\usepackage{multirow}
\usepackage{authblk}

\author[1,2]{Gui Wang \textsuperscript{\dag}}
\author[1]{Yang Wennuo \textsuperscript{\dag}}
\author[1]{Xusen Ma}
\author[1]{Zehao Zhong}
\author[1]{Zhuoru Wu}
\author[4]{Ende Wu}
\author[3]{Rong Qu}
\author[2]{Wooi Ping Cheah}
\author[2]{Jianfeng Ren\textsuperscript{*}}
\author[1]{Linlin Shen\textsuperscript{*}}

\affil[1]{School of Computer Science \& Software Engineering, Shenzhen University, Shenzhen, China}
\affil[2]{School of Computer Science, University of Nottingham Ningbo China, Ningbo, China}
\affil[3]{School of Computer Science, University of Nottingham, Nottingham, United Kingdom}
\affil[4]{Wenzhou Medical University, Wenzhou, China}
\affil[ ]{\textsuperscript{*}Corresponding authors: \texttt{llshen@szu.edu.cn}}

\def \eg {\textit{e.g.}}
\def \ie {\textit{i.e.}}

\title{EyePCR: A Comprehensive Benchmark for Fine-Grained Perception, Knowledge Comprehension and Clinical Reasoning in Ophthalmic Surgery}

\begin{document}

\maketitle

\begin{abstract}
MLLMs (Multimodal Large Language Models) have showcased remarkable capabilities, but their performance in high-stakes, domain-specific scenarios like surgical settings, remains largely under-explored. To address this gap, we develop \textbf{EyePCR}, a large-scale benchmark for ophthalmic surgery analysis, grounded in structured clinical knowledge to evaluate cognition across \textit{Perception}, \textit{Comprehension} and \textit{Reasoning}. EyePCR offers a richly annotated corpus with more than 210k VQAs, which cover 1048 fine-grained attributes for multi-view perception, medical knowledge graph of more than 25k triplets for comprehension, and four clinically grounded reasoning tasks. The rich annotations facilitate in-depth cognitive analysis, simulating how surgeons perceive visual cues and combine them with domain knowledge to make decisions, thus greatly improving models' cognitive ability. In particular, \textbf{EyePCR-MLLM}, a domain-adapted variant of Qwen2.5-VL-7B, achieves the highest accuracy on MCQs for \textit{Perception} among compared models and outperforms open-source models in \textit{Comprehension} and \textit{Reasoning}, rivalling commercial models like GPT-4.1. EyePCR reveals the limitations of existing MLLMs in surgical cognition and lays the foundation for benchmarking and enhancing clinical reliability of surgical video understanding models. 
\end{abstract}


\section{Introduction}
Multimodal large language models such as GPT-4o~\cite{openai2024gpt4o}, Gemini~\cite{team2023gemini} and LLaVA~\cite{liu2023visual} have demonstrated significant capabilities in visual question answering (VQA), instruction following and cross-modal reasoning. However, their performance in scenarios that require domain-specific visual perception and structured semantic understanding, especially high-stakes medical scenarios, remains under-explored~\cite{twinanda2016endonet,jin2020multi}. Surgical video analysis poses unique challenges, including real-time procedures in constrained anatomical space, subtle yet consequential actions, and highly demanding interpretation on not only what is seen but also why it is done, what comes next, and what risks may arise~\cite{ma2021self,gao2022transsvnet}. 

Previous surgical video analysis focuses on low-level tasks like tool segmentation~\cite{Cataract-1k, SurVLP}, phase recognition~\cite{hu2024ophnet, gupta2023dataset}, and temporal action detection~\cite{ophclip,Gp-vls}. While valuable, these tasks do not probe models' capacity on multimodal cues fusion, domain knowledge alignment, or reasoning over surgical intent, procedural risks and intraoperative decision paths. Recent datasets now focus on surgical VQAs~\cite{TEMSET-24K, Surg-QA, gupta2023dataset}, 
but they overuse large language models (LLMs) to generate VQAs from semantics/captions, causing hallucinations and poor surgical grounding~\cite{chen2024multi, kim2024code, zhang2024automated}.
Consequently, it cannot guarantee the factual accuracy or clinical validity. Ophthalmic surgery presents unique challenges, \eg, fine-grained instrument handling, microanatomical variations, and high sensitivity to visual cues like anterior chamber depth, capsule tears, and corneal edema, all of which demand structured, knowledge-informed comprehension. 
\begin{figure}[!t]
\centering
\includegraphics[width=0.9\textwidth]{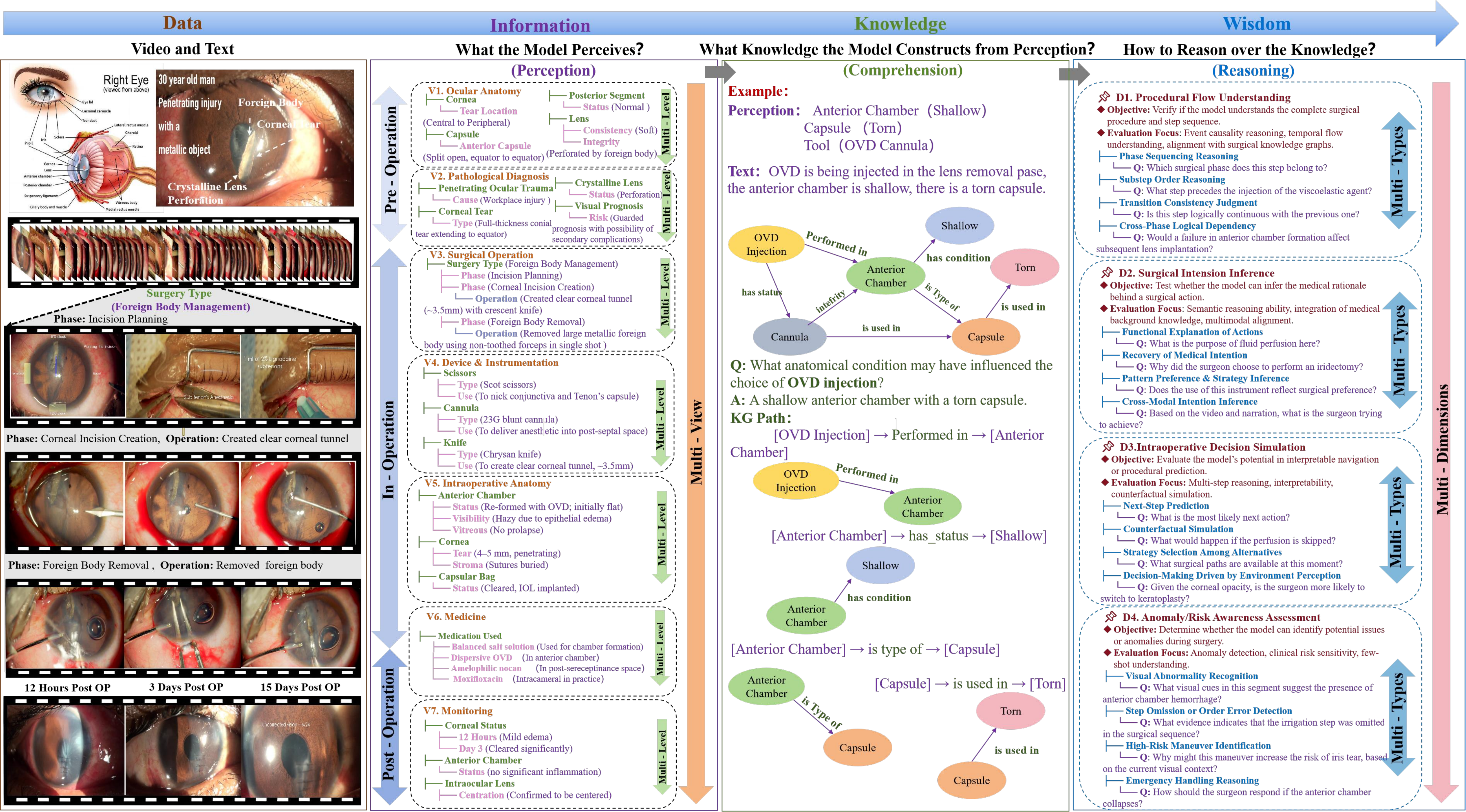}
\caption{Overview of the EyePCR framework, consisting of a three-stage cognitive hierarchy: fine-grained perception, scene graph comprehension, and multi-dimensional clinical reasoning.  
}
\label{overview}
\end{figure}

To bridge the gap, we develop \textbf{EyePCR}, a novel benchmark for evaluating surgical cognitive abilities under a \textbf{PCR} (\textbf{Perception $\rightarrow$ Comprehension $\rightarrow$ Reasoning}) paradigm shown in Fig.~\ref{overview}.  
For \textbf{Perception}, the model extracts \textit{multi-view, multi-level attributes} from videos.  
The fine-grained annotations are curated across diverse semantic perspectives, \eg, anatomy, tools, procedures, and physiological states, simulating how experts observe and interpret the operative field. 
For \textbf{Comprehension}, sensory observations are integrated into \textit{Structured Scene Graphs} to align low-level cues with high-level surgical knowledge, each encoding intraoperative semantics using \textit{entity–action–target} triplets grounded in surgical ontologies, \eg, [\texttt{Forceps}] $\rightarrow$ \texttt{grasp} $\rightarrow$ [\texttt{Capsular Flap}]. Scene graphs are temporally aggregated into a \textit{Semantic Memory Graph}, preserving procedural context and aligning observed events with clinical intent. 
Our structured graphs support a suite of reasoning tasks that are \textit{knowledge grounded}, \textit{clinically valid} and \textit{interpretable}. 
For \textbf{Reasoning}, four tasks are defined: \textit{Procedural Flow Understanding} to verify the understanding over the complete surgical procedure; \textit{Surgical Intent Inference} to assess medical reasoning inference for surgical actions; 
\textit{Intraoperative Decision Simulation} to assess interpretable navigation/procedural prediction potential;
and \textit{Anomaly/Risk Awareness Assessment} to evaluate surgical anomaly identification ability. Beyond visual perception, these tasks interpret, predict and reason over surgical behaviours using knowledge graphs.

EyePCR is the first large-scale (210K) ophthalmic surgery VQA corpus grounded in structured clinical knowledge, with detailed annotations across all three cognitive stages: 
1)~102K VQAs annotated with 1048 multi-view, multilevel attributes for fine-grained perception; 
2)~24K VQAs supported by a scene graph abstraction of 25K triplets, providing structured knowledge for comprehension; 
3)~86K VQAs spanning four reasoning tasks, each evaluating a distinct dimension of surgical intelligence.

Our main contributions are three-fold: 
1)~The proposed PCR framework offers a unified paradigm for evaluating surgical cognitive models by jointly integrating fine-grained visual perception, structured knowledge comprehension, and high-level clinical reasoning, supporting the construction of \textit{knowledge-grounded}, \textit{interpretable}, and \textit{clinically reliable} surgical cognition models. 
2)~The \textbf{EyePCR} dataset benchmarks MLLMs in real-world medical decision-making, moving from what models can \textit{see}, to what they can \textit{comprehend} and \textit{decide}.
3) The PCR paradigm could greatly enhance the cognitive ability, \ie, \textbf{EyePCR-MLLM}, a domain-adapted variant of Qwen2.5-VL-7B, outperforms open-source multimodal models and performs on par with commercial models like GPT-4.1~\cite{openai2024gpt41}.

\section{Review of Surgical Video Analysis Datasets} 
For surgical videos, most datasets focus on perception~\cite{SurVLP,ophclip, hu2024ophnet, li2023llava_med}, 
while recent ones focus more on understanding/reasoning \cite{SSG-VQA, gupta2023dataset, TEMSET-24K}. 
Table~\ref{tab:eyeqa-comparison} summarizes surgical video analysis datasets. Only EyePCR supports full-spectrum annotations for perception, comprehension and reasoning. \quad \quad \quad 
\begin{table}[!t]
\centering
\caption{
Dataset comparison for surgical video analysis. 
\textbf{EyePCR} is the only dataset that supports full-spectrum annotations from multi-view perception (\textbf{V1–V7} defined in Sec.~\ref{sec:perception}) to knowledge graph (\textbf{KG}) comprehension and reasoning across four clinical dimensions (\textbf{D1–D4} defined in Sec.~\ref{sec:reasoning}). 
}
\label{tab:eyeqa-comparison}
\resizebox{\textwidth}{!}{
\begin{tabular}{lccrcccccccc|c|ccccc}
\toprule
\multirow{2}{*}{\textbf{Dataset}} & \multirow{2}{*}{\textbf{Year}} & \multirow{2}{*}{\textbf{Domain}} & \multirow{2}{*}{\textbf{Size}} & \multirow{2}{*}{\textbf{Data}} 
& \multicolumn{7}{c|}{\textbf{P}} 
& \textbf{C} & \multicolumn{4}{|c}{\textbf{R}} \\
\cmidrule(lr){6-12} \cmidrule(lr){13-13} \cmidrule(lr){14-17}
& & & & & V1 & V2 & V3 & V4 & V5 & V6 & V7 & KG & D1 & D2 & D3 & D4 \\
\midrule
Cholecseg~\cite{Cholecseg8k}    & 2020 & Laparoscopic & 8K     & Object-Label & \xmark & \xmark & \xmark & \cmark & \cmark & \xmark & \xmark & \xmark & \xmark & \xmark & \xmark & \xmark \\ \hline
SurVLP~\cite{SurVLP}            & 2023 & Endoscope    & 25K    & Video-Label & \xmark & \xmark & \cmark & \cmark & \xmark & \xmark & \xmark & \xmark & \xmark & \xmark & \xmark & \xmark \\
OphNet~\cite{hu2024ophnet}      & 2024 & Ophthalmic   & 14K    & Video-Label & \xmark & \xmark & \cmark & \xmark & \xmark & \xmark & \xmark & \xmark & \xmark & \xmark & \xmark & \xmark \\
Cataract~\cite{Cataract-1k}     & 2024 & Ophthalmic   & 1K     & Video-Label & \xmark & \xmark & \cmark & \cmark & \cmark & \xmark & \xmark & \xmark & \xmark & \xmark & \xmark & \xmark \\
AVOS~\cite{AVOS}                & 2024 & Generalist   & 2K   & Video-Label & \xmark & \xmark & \xmark & \cmark & \xmark & \cmark & \xmark & \xmark & \xmark & \xmark & \xmark & \xmark \\
MedVidQA~\cite{gupta2023dataset} & 2024 & Aid        & 3.4K   & Video-Label & \xmark & \xmark & \xmark & \xmark & \xmark & \xmark & \xmark & \xmark & \xmark & \xmark & \cmark & \xmark \\
TEMSET~\cite{TEMSET-24K}        & 2025 & Endoscope    & 24K    & Video-Label & \xmark & \xmark & \cmark & \xmark & \xmark & \xmark & \xmark & \xmark & \xmark & \xmark & \cmark & \xmark \\ \hline
GenSurg+~\cite{gensurg}         & 2024 & Laparoscopic & 17K    & Video-Caption  & \xmark & \xmark & \cmark & \xmark & \xmark & \xmark & \xmark & \xmark & \xmark & \xmark & \xmark & \xmark \\
HecVL~\cite{Hecvl}              & 2024 & Endoscope    & 37K    & Video-Caption  & \xmark & \xmark & \cmark & \xmark & \xmark & \xmark & \xmark & \xmark & \xmark & \xmark & \xmark & \xmark \\
OphVL~\cite{ophclip}            & 2025 & Ophthalmic   & 375K   & Video-Caption  & \xmark & \xmark & \cmark & \cmark & \xmark & \cmark & \xmark & \xmark & \xmark & \xmark & \xmark & \xmark \\ \hline
GP-VLS~\cite{Gp-vls}            & 2024 & Generalist   & 120K   & VQA   & \xmark & \xmark & \cmark & \cmark & \xmark & \xmark & \xmark & \xmark & \xmark & \xmark & \xmark & \xmark \\
SSG-VQA~\cite{SSG-VQA}                & 2024 & Generalist   & 960K   & VQA & \xmark & \xmark & \xmark & \cmark & \cmark & \xmark & \xmark & \cmark & \xmark & \cmark & \xmark & \xmark \\
Surg-QA~\cite{Surg-QA}          & 2025 & Generalist   & 102K   & VQA   & \xmark & \xmark & \cmark & \cmark & \xmark & \xmark & \xmark & \xmark & \cmark & \xmark & \cmark & \xmark \\
\rowcolor{gray!30}
\textbf{EyePCR(Ours)}          & \textbf{2025} & \textbf{Ophthalmic} & \textbf{210K} & \textbf{VQA} & 
\cmark & \cmark & \cmark & \cmark & \cmark & \cmark & \cmark & \cmark & \cmark & \cmark & \cmark & \cmark \\
\bottomrule
\end{tabular}
}
\end{table}
\noindent\textbf{Fine-Grained Perception.} \quad 
Existing surgical datasets use coarse labels like phase boundaries~\cite{twinanda2016endonet, jin2020multi} or action clips~\cite{lea2016temporal}, restricting fine-grained visual understanding and semantic analysis. 
Datasets like Cataract~\cite{schoeffmann2018cataract} and OphNet~\cite{hu2024ophnet} include frame-/segment-level annotations but lack structured information on surgical tools, anatomy states, or procedural context, restricting downstream fine-grained visual reasoning tasks. 
In contrast, EyePCR adopts a hierarchy to annotate 1048 fine-grained attributes across seven dimensions (V1–V7) as defined in Sec.~\ref{sec:perception}.  
While fine-grained video understanding has been widely explored in activity recognition~\cite{piergiovanni2020fine}, instructional video modeling~\cite{miech2019howto100m}, and object–action co-localization~\cite{belharbi2023colo}, surgical video analysis presents unique challenges in constrained perceptive environments, task-specific instruments, and high annotation costs. 
EyePCR bridges the gap by offering a large-scale, clinically validated dataset with rich fine-grained annotations, laying a foundation for multimodal reasoning in surgical AI.

\noindent\textbf{KG-based Comprehension.} \quad 
While knowledge graphs (KGs) enable multi-hop reasoning and semantic alignment in natural image understanding~\cite{li2020boosting, zellers2018neural} and VQAs~\cite{wang2022vqa, teney2017graph}, they remain unexplored for surgical video analysis. Existing surgical datasets like SurVLP~\cite{SurVLP}, HecVL~\cite{Hecvl}, and MedVidQA~\cite{gupta2023dataset} lack entity–relation modelling and semantic scene representations, weakening the support for temporal reasoning, causal chaining, and inter-step coherence across surgical workflows. 
EyePCR uniquely constructs knowledge-anchored surgical scene graphs from fine-grained attributes, bridging multimodal cues to high-level reasoning. 
Compared to prior KGs~\cite{Radgraph,BioGPT}, EyePCR uniquely aligns clinical knowledge with dynamic, video-based surgical environments, grounded in real-time visual observations, structured procedural flow, and clinically meaningful task sequences. 

\noindent\textbf{Surgical Video Reasoning.} \quad 
Surgical video reasoning grows with multimodal VQA datasets like MedVidQA~\cite{gupta2023dataset}, Surg-QA~\cite{Surg-QA}, and SSG-VQA~\cite{SSG-VQA}. But these datasets emphasize general-purpose/retrieval-style shallow VQAs on tool identification, anatomy naming, and isolated action recognition, more towards visual recognition than surgical cognition. 
EyePCR explicitly aligns its QA structure with four surgical reasoning dimensions, mirroring surgeons' contextual and temporal decision-making. 
Compared to medical VQA datasets centred on diagnostic image interpretation like VQA-RAD~\cite{lau2018dataset}, PathVQA~\cite{he2020pathvqa}, and SLAKE~\cite{liu2021slake}, EyePCR tackles surgical reasoning's core challenges, establishing the first benchmark combining multimodal QA with clinical reasoning.

\section{EyePCR Dataset Construction}
\subsection{Data Collection and Curation} 
\noindent \textbf{Video Collection and Preprocessing.} \quad
We collected a total of 2,968 surgical videos from public datasets like OphNet~\cite{hu2024ophnet}, LensID~\cite{LensID}, and Cataract-1k~\cite{Cataract-1k}, and from curated clinical repositories~\cite{willsrepo2024}, selected institutional archives~\cite{wmu_repo2024}, Eyetube~\cite{eyetube} and American Academy of Ophthalmology~\cite{aao}. Personally identifiable information is removed from videos. All involved videos are publicly sourced.
Each video underwent preliminary preprocessing 
to ensure procedural completeness, 
and the presence of narration or embedded subtitles to support textual information extraction. 

\noindent\textbf{Expert Screening.} \quad
Three senior ophthalmic surgeons independently reviewed all the 2,968 videos based on four criteria: 
1)~Demonstrating procedural completeness, covering essential surgical phases without critical omissions; 
2)~Maintaining educational clarity, with a well-illuminated, unobstructed, and centered surgical field, minimizing instrument-induced occlusions or recording artifacts; 
3)~Exhibiting clinical representativeness, depicting standard techniques across various ophthalmic procedures;  
4)~Presenting consistent and informative narration or subtitles to accurately describe key surgical manoeuvrers and clinical decision points. 
Following the consensus of experts, a curated subset of 1,544 high-quality videos was retained, forming a solid foundation for subsequent structured annotations. The expert screening ensures clinical authenticity and reliability of the EyePCR dataset. 

\noindent \textbf{Textual Information Extraction.} \quad 
A dual-mode textual extraction strategy was employed to extract accurate procedural narration.
1)~Automated speech recognition was performed using OpenAI’s Whisper~\cite{whisper2022} to generate high-fidelity transcriptions of narrated surgical procedures. 
2)~Optical character recognition (OCR) was conducted using CnOCR~\cite{cnocr2023}, a multilingual OCR toolkit, to extract embedded subtitle texts from video frames. Domain-specific noise reduction heuristics~\cite{liu2021syntactic} were applied to ensure textual consistency and fidelity. 
The generated text corpus serves as a reliable basis for structured knowledge extraction, scene graph construction, and VQA generation. 

\subsection{Perception Hierarchy of Multi-View Multi-Level Attributes}
\label{sec:perception}
\begin{figure}[!b]
\centering
\includegraphics[width=0.8\textwidth]{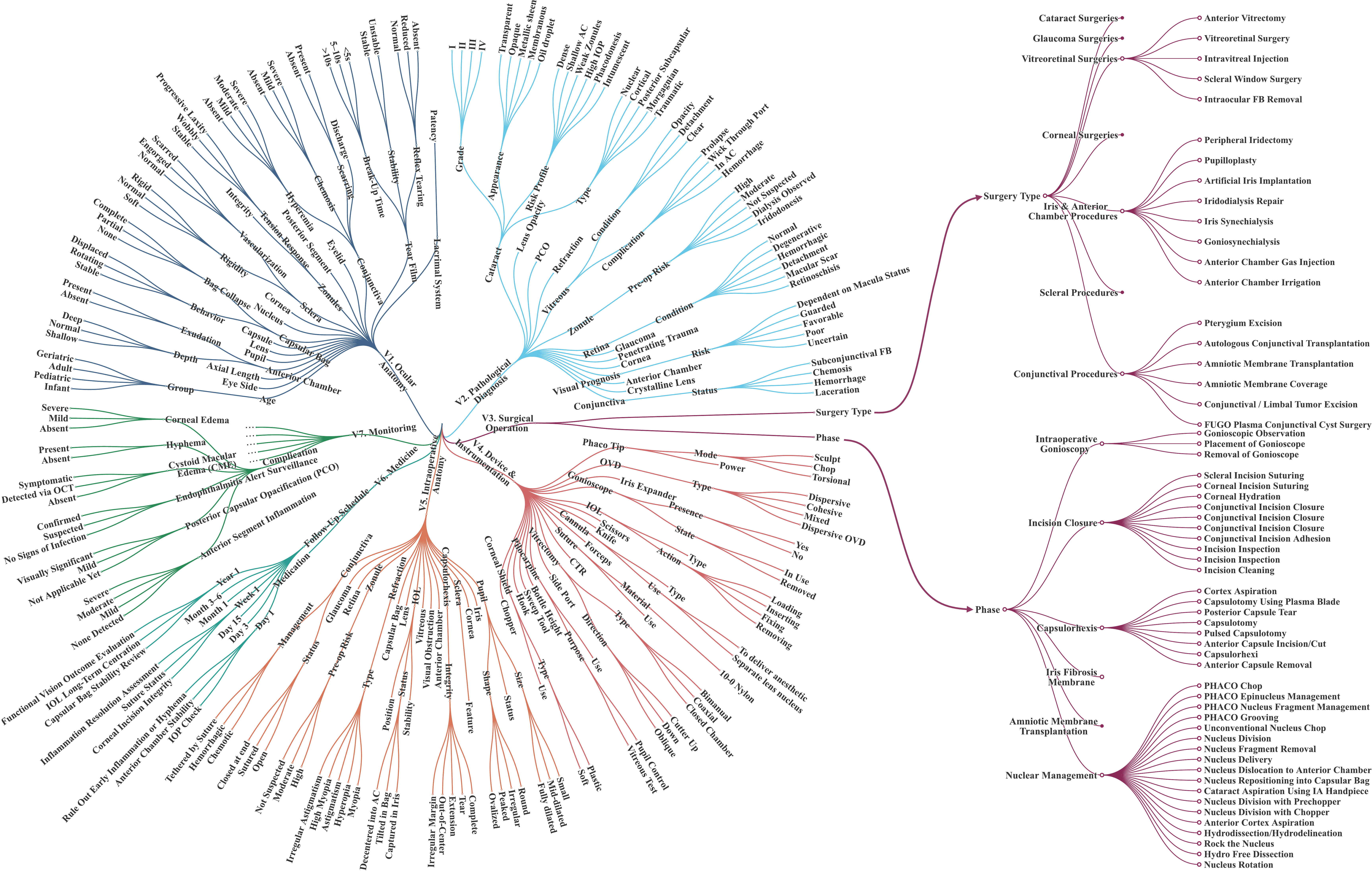}
\caption{Hierarchy of multi-view multi-level fine-grained attributes on the full surgical continuum.}  
\label{fig2}
\end{figure}
\noindent\textbf{Multi-View Attributes.} \quad 
Fig.~\ref{fig2} depicts seven semantic views to capture distinct yet interrelated facets of surgical cognition.  
1)~\textit{Ocular Anatomy}, encoding anatomical baselines essential for assessing the physiological state of the eye before intervention; 
2)~\textit{Pathological Diagnosis}, documenting lesion characteristics, trauma indicators and pathological risks; 
3)~\textit{Surgical Operation}, capturing procedural sequences and operational intent at different phases; 
4)~\textit{Device \& Instrumentation}, tracking the presence, type and usage state of surgical tools; 
5)~\textit{Intraoperative Anatomy}, reflecting real-time changes in ocular structures during manipulation; 
6)~\textit{Medicine}, recording pharmacological management; 
7)~\textit{Monitoring}, ensuring ongoing anatomical and functional surveillance post-surgery.
Together, these views enable a multidimensional reconstruction of the surgical context, supporting visual reasoning grounded in real-world clinical scenarios. 

\noindent\textbf{Multi-Level Attributes} \quad
Surgical attributes are organized into three levels, reflecting how surgeon progressively process visual and procedural details during surgery. 
1)~\textit{L1} outlines broad anatomical or procedural domains such as cornea, anterior chamber and surgical tools;  
2)~\textit{L2} refines these into specific sub-features like corneal clarity or iris stability, capturing clinically relevant cues;  
3)~\textit{L3} provides fine-grained descriptions like shape, location and dynamic state. 
The layered structure enables LLMs to perceive, organize, and reason with surgical data across multiple views and levels, extracting clinically meaningful insights with clarity, coherence, and purpose throughout the surgery.


\noindent \textbf{Statistics of Annotated Attributes.} \quad 
Each video is segmented into temporal units, which are then annotated with fine-grained attributes. 
In total, 1,048 fine-grained attributes are annotated across seven views at three levels, \ie, 76 \textit{L1} coarse attributes, 240 \textit{L2} subtypes, and 732 \textit{L3} detailed attributes. 
The multi-view multi-level annotations ensure that surgical scenes are perceived at various granularity, providing a rich perceptual substrate for downstream understanding and reasoning tasks. 

\subsection{Graph-based Comprehension of Surgical Knowledge}
\noindent\textbf{Structured Scene Graph} \quad 
To move beyond low level perception, we transform diverse visual attributes into coherent semantic graph representations, offering a structured foundation for clinically grounded comprehension. Specifically, for each video segment, we extract visual annotations in seven semantic views, and process video subtitles to align textual descriptions with visual observations. We then construct a \textit{Structured Scene Graph} from these two sources of information. Each graph comprises a set of \textit{entity–action-attribute} triplets that represent surgical interpretable events, \eg, \texttt{[OVD]} $\rightarrow$ \texttt{injected into} $\rightarrow$ \texttt{anterior chamber}, allowing each predicted answer to be directly traced back to structured visual-semantic evidence grounded in the video. 

\noindent\textbf{Semantic Memory Graph} \quad 
To capture temporal and contextual continuity, we aggregate the segment-level scene graphs across all videos into a \textit{Semantic Memory Graph}, recording the evolution of the surgery by linking related entities and actions over time, modelling procedural flow, step dependencies, and recurrent structures, \eg, repeated use of a device or transitions between phases. 
The semantic memory graph encodes spatial structure of the surgery in a clinically meaningful way, and encodes temporal continuity across video segments, enabling the model to perform cross-step reasoning and causal inference such as understanding action preconditions or predicting subsequent maneuvers, thereby preserving the procedural integrity of surgical workflows. 

\noindent\textbf{Reasoning Paths} \quad 
We extract \textit{Reasoning Paths} from the semantic memory graph to support high-level reasoning tasks. Each path is a multi-hop subgraph capturing clinically meaningful event sequences, such as \texttt{incision} $\rightarrow$ \texttt{viscoelastic injection} $\rightarrow$ \texttt{capsulorhexis}. These paths simulate how surgeons reason through procedural steps and their dependencies. To construct reasoning-focused VQA questions, we use these paths as templates. For example, a path ending in \texttt{capsulorhexis} may yield the question “What steps are typically required before capsulorhexis?” or “Why is viscoelastic injected at this stage?” This ensures that each question is grounded in verifiable surgical logic, enabling interpretable and clinically aligned model evaluation. These reasoning paths not only serve as templates for constructing clinically meaningful VQAs, but also constrain model outputs to medically valid and contextually grounded event sequences, mitigating hallucinations and enabling interpretable, logic-driven surgical reasoning. 


\noindent\textbf{Statistics of Knowledge Graph} \quad
The constructed semantic memory graph contains a total of 25,567 triplets, covering 16,034 entities and 3,507 relations. We further extract 11,462 logical paths from the knowledge graph, including 9,499 two-hop paths and 1,963 paths involving three or more hops. 


\subsection{Surgical Video Reasoning}
\label{sec:reasoning} 
While perception and comprehension focus on extracting and structuring visual semantics, reasoning requires a model to interpret, infer, and simulate surgical cognition. We define four medically grounded evaluation dimensions.  
1)~\textbf{Procedural Flow Understanding} targets the model’s capacity to understand temporal and causal structure of surgical workflows, \eg, identifying sub-step ordering within a phase, recognizing cause–effect relationships between adjacent or temporally distant actions, and verifying clinically valid logic of transitions across phases. 
Such tasks emphasize not just recognition of step labels, but also grasp of procedural flow and inter-step dependencies. 
2)~\textbf{Surgical Intent Inference} evaluates the ability of correctly interpreting the clinical rationale behind a given action, \eg, attributing actions to goals, 
recognizing patterns in strategy selection, 
and aligning procedural behaviours with physiological or pathological conditions present in the operative field. 
In particular, rather than identifying tools or anatomy in isolation, the model must infer why a specific tool is used at a particular stage, and how the usage aligns with the surgical objective. 
3)~\textbf{Intraoperative Decision Simulation} simulates real-time surgical planning and adaptive decision-making, \eg, forecasting the next likely step in an ongoing procedure, planning contingencies for complications like anterior chamber collapse, proposing viable alternatives under uncertainty, and adapting its response based on observed scene conditions. 
These tasks evaluate the ability to generalize beyond static recognition and perform reasoning under dynamic constraints. 
4)~\textbf{Anomaly/Risk Awareness Assessment} focuses on the ability to identify deviations from expected patterns and infer their clinical implications, \eg, detecting subtle pathological changes, 
predicting complications based on anatomical or procedural abnormalities, and assessing the potential consequences of incorrect actions. These tasks require integrating visual cues with latent medical knowledge to simulate expert-level risk sensitivity and procedural safety judgment. 
Each dimension contains multiple high-level clinical inference types, ranging from temporal logic modelling, surgical goal attribution, strategic forecasting to risk simulation. These tasks are instantiated as VQAs, grounded in structured scene graphs and domain ontologies, enabling a rigorous and interpretable evaluation of surgical cognition. 

\subsection{Construction of VQAs for Perception, Comprehension and Reasoning} 
\begin{figure}[!t]
\centering
\includegraphics[width=\textwidth]{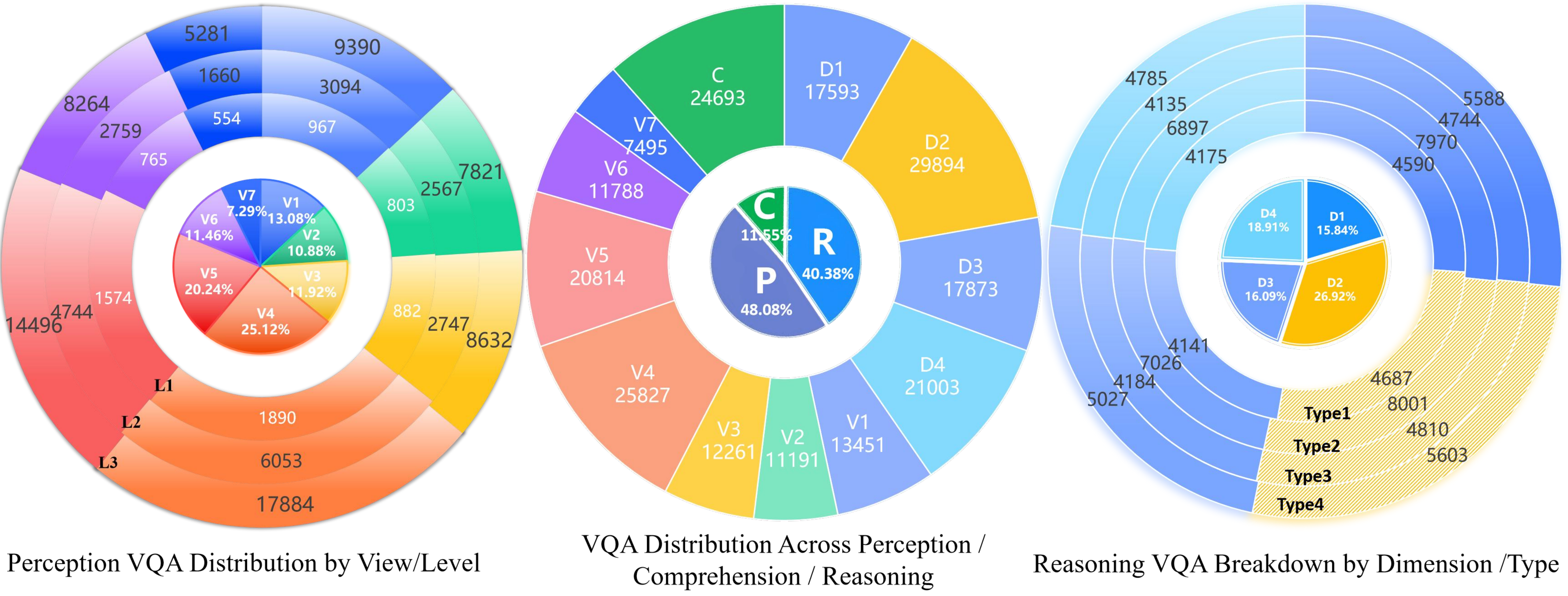}
\caption{Distribution of 210K VQAs across \textbf{Perception}, \textbf{Comprehension}, and \textbf{Reasoning} stages. 
}
\label{VQA-Sta}
\end{figure}

\noindent \textbf{VQA on Attributes} \quad  
We design a comprehensive set of VQA templates that cover observable surgical scene properties, which are readily identifiable from video segments. 
By using \texttt{Gemini-2.5-Pro}~\cite{google2024gemini}, we automatically generate structured VQA pairs for each video segment, where two types of questions are created: multiple-choice questions (MCQs) like \textit{``Which ocular structure is being manipulated in this segment?''} with options \textit{\{``Cornea'', ``Conjunctiva'', ``Sclera'', ``Iris''\}}, and open-ended questions (OEQs) like \textit{``Describe the baseline anatomical condition of the cornea prior to any surgical intervention''}.
As shown in the left panel of Fig.~\ref{VQA-Sta}, we generate \textbf{64,426 MCQs} and \textbf{38,401 OEQs} from \textbf{47,336 video segments}, across seven semantic views under the perception stage. These questions are further stratified across three attribute levels, ensuring both breadth and depth of perceptual VQAs. This stage focuses on evaluating the model's ability to parse diverse, multi-layered clinical attributes directly from surgical video segments. 

\noindent\textbf{VQA on KG-based Comprehension.}  \quad
VQAs in the comprehension stage are designed to evaluate whether models can go beyond perception and interpret structured surgical semantics grounded in knowledge graphs. To this end, we generate \textbf{24,693} OEQs from \textbf{16,556 video segments} based on \textit{entity–action-attribute} triplets extracted from scene graphs, capturing intraoperative actions, object interactions, and contextual relations, \eg, \textit{``Describe the path of the anchoring suture''}, where the expected answer would be \textit{``Passes into the sclera near the limbus and out through the edge of the conjunctival graft''}, following the reasoning path: 
\texttt{[Suture]} $\rightarrow$ \texttt{passes into} $\rightarrow$ \texttt{Sclera} and \texttt{[Suture]} $\rightarrow$ \texttt{related to} $\rightarrow$ \texttt{Conjunctival Graft}.
As shown in the middle panel of Fig.~\ref{VQA-Sta}, this stage accounts for approximately 11.5\% of the total VQAs and serves as a bridge between low-level visual parsing and high-level surgical reasoning, emphasizing more on intra-segment comprehension. 

\noindent\textbf{VQA on Surgical Reasoning.}  \quad
We construct \textbf{86,363} OEQs from \textbf{19,001 video segments} grounded on reasoning paths, targeting four core reasoning dimensions. 
For \textit{Procedural Flow Comprehension}, questions like \textit{``Which surgical phase does this action belong to?''} assess the understanding on stepwise dependencies across stages; 
For \textit{Surgical Intent Inference}, questions like \textit{``Why did the surgeon choose to perform an iridectomy?''} assess action purpose inference from contextual cues; 
For \textit{Intraoperative Decision Simulation}, questions like \textit{``What would happen if the perfusion is skipped?''} require the model to simulate clinical decision-making; 
For \textit{Anomaly/Risk Awareness Assessment}, questions like \textit{``What potential complications may arise from this action involving the iris?''} 
simulate risk-aware behaviour. 
Open-ended questions reflect surgical reasoning complexity, avoiding fixed-option limitations and accommodating plausible alternatives. 

\noindent\textbf{Statistics of VQAs}  \quad
We curate a total of 82,893 video segments from 1,544 full-length ophthalmic surgery videos, generating 213,883 high-quality VQAs, \ie, 64,425 MCQs and 149,458 OEQs, 
categorized into three cognitive stages: \textbf{102,827} for \textbf{Perception}, \textbf{24,693} for \textbf{Comprehension}, and \textbf{86,363} for \textbf{Reasoning}. The VQAs are distributed as shown in Fig.\ref{VQA-Sta}, reflecting our emphasis from low-level visual perception, middle-level comprehension to high-level clinical reasoning. 

\subsection{Evaluation Protocols}

\noindent\textbf{Dataset Split.} \quad 
The EyePCR dataset is split into training and testing sets at the patient level, preventing any overlap of patient identity or videos across partitions. 
The dataset contains a total of 82,893 annotated video segements, of which 77,133 are allocated to training and 5,760 to testing, 
along with 198,630 VQAs for training and 15,253 VQAs for testing. The training set includes 95,246 VQAs for perception, 23,072 for  comprehension, and 80,312 for reasoning, while the test set consists of 7,581, 1,621, and 6,051 VQAs across the three categories. 
This structured split enables a fine-grained assessment of model capabilities across perception, comprehension, and reasoning tasks, ensuring fair and reproducible performance comparisons. 

\noindent\textbf{Evaluation Metrics.} \quad 
For MCQs, simple accuracy is used. For OEQs, we employ: 1) A suite of linguistic metrics, including BLEU-1 through BLEU-4 and ROUGE-1~\cite{papineni2002bleu}, ROUGE-2, and ROUGE-L~\cite{lin2004rouge}, to evaluate lexical overlap with ground-truth answers; 2) GPT-4.1-based rubric scoring~\cite{openai2024gpt4o, liu2023gpt4rubric}, rated on a 5-point Likert scale for factual correctness, specificity, and completeness, by prompting GPT-4.1 with standardized grading instructions to evaluate clinical relevance of all model-generated responses across the entire OEQ test set.


\section{Experimental Results}
\label{sec:experimental_results} 
\subsection{Experimental Setup}

\noindent \textbf{Evaluation Models.} \quad 
The proposed EyePCR benchmark is evaluated on a broad spectrum of MLLMs, encompassing both commercial and open-source systems across varying capability tiers. 
\textbf{Commercial models} include \textbf{GPT-4.1}~\cite{openai2024gpt41}, \textbf{GPT-o3}~\cite{openai2024gpt3.5}, \textbf{Claude 3.7 (Sonnet)}~\cite{claude2024sonnet}, as well as \textbf{Gemini-2.0-Flash}, \textbf{Gemini-2.5-Flash}, and \textbf{Gemini-2.5-Pro}~\cite{google2024gemini}. These models represent the cutting edge in general-purpose multimodal reasoning, with strong performance across vision-language and video-based tasks. 
\textbf{Open-source models} include \textbf{Qwen2.5-VL-7B}, \textbf{Qwen2.5-VL-32B}~\cite{Qwen2.5-VL}, \textbf{InternVL-14B}~\cite{internvl2024}, \textbf{LLaVA-Video-7B}~\cite{LLaVA-Video-7B}, and \textbf{VideoLLaMA3-7B}~\cite{VideoLLaMA3-7B}. 
In addition, we fine-tune the open-source \textbf{Qwen2.5-VL-7B}~\cite{Qwen2.5-VL} on the OpenEye training dataset, yielding a domain-specialized variant, \textbf{EyePCR-MLLM}, tailored for surgical video understanding and clinical VQA tasks. The fine-tuning is performed across all three cognitive levels, enabling the model to internalize structured surgical knowledge and improve multimodal alignment. Training details are given in Appendix. 

Beside, we incorporate human experts to contextualize model performance. We invited \textbf{66 ophthalmology professionals}, including \textbf{47 Ophthalmology Trainees} and \textbf{19 Ophthalmic Surgeons}, to participate in the evaluation. Each expert was randomly assigned one of five questionnaires, with a total of \textbf{50 MCQs} and \textbf{20 OEQs}, covering all three cognitive stages, \textbf{Perception}, \textbf{Comprehension}, and \textbf{Reasoning}, across diverse semantic views. This human evaluation provides a reference point for assessing both perceptual accuracy and clinical reasoning capacity, and highlights the performance gap between domain-informed human experts and general-purpose MLLMs.

\subsection{Evaluation on MCQs for Fine-Grained Perception} 
Tab.~\ref{tab:MCQ} summarizes the MCQ comparisons 
for fine-grained perception. We can observe the following.  
\begin{table*}[!t]
\centering
\caption{
\textbf{MCQs} performance across seven perceptual views (V1–V7) in the \textbf{Perception} stage. 
\textcolor{blue}{\textbf{Blue}}: highest among \textbf{commercial models}; 
\textcolor{red}{\textbf{Red}}: highest among \textbf{open-source models}; 
\textbf{Bold}: best overall.} 
\label{tab:perception-mcq-accuracy}
\small
\vspace{0.5em}
\renewcommand{\arraystretch}{1.3}
\resizebox{\textwidth}{!}{
\begin{tabular}{cl|ccccccc|c}
\toprule
\textbf{} & \textbf{Participants} & \textbf{V1} & \textbf{V2} & \textbf{V3} & \textbf{V4} & \textbf{V5} & \textbf{V6} & \textbf{V7} & \textbf{Avg} \\
\midrule
\multirow{6}{*}{\rotatebox[origin=c]{90}{\textbf{Commercial}}} 
& Gemini 2.0 Flash~\cite{google2024gemini}       & 0.6688 & 0.7144     & 0.5144 & 0.6810 & 0.7344   & 0.4946 & 0.5409 & 0.6212 \\
& Claude 3.7~\cite{claude2024sonnet}       & 0.7436 & 0.7932   & 0.5108 & 0.6324 & 0.7853   & 0.4864 & 0.6191 & 0.6530 \\
& Gemini 2.5 Flash~\cite{google2024gemini} & 0.7173 & 0.7744     & 0.5974 & 0.6743 & 0.7673   & 0.5055 & 0.6170 & 0.6647 \\
& GPT-o3~\cite{openai2024gpt3.5}           & 0.7172 & 0.7849     & \textcolor{blue}{0.6555} & 0.7080 & 0.7835   & 0.5209 & 0.6710 & 0.6916 \\
& GPT-4.1~\cite{openai2024gpt41}            & \textcolor{blue}{0.7859} &   0.8432  & 0.6360 & \textcolor{blue}{0.7157} & 0.8012   & 0.5258 & 0.6407 & 0.7069\\
& Gemini 2.5 Pro~\cite{google2024gemini}   & 0.7505 & \textcolor{blue}{0.8514}     & 0.6530 & 0.6982 & \textcolor{blue}{0.8334}   & \textcolor{blue}{0.5553} & \textcolor{blue}{0.6551} & \textcolor{blue}{0.7138} \\
\midrule
\multirow{6}{*}{\rotatebox[origin=c]{90}{\textbf{Open-Source}}} 
& VideoLLaMA3-7B~\cite{VideoLLaMA3-7B}   & 0.2098     &  0.2660    &  0.1270    &   0.3554   &  0.3124    &  0.1987    & 0.1450     & 0.2158 \\
& LLaVA-Video-7B~\cite{LLaVA-Video-7B}     & 0.2886 & 0.2960     & 0.2009 & 0.2148  & 2614   & 0.2094 & 0.1947 & 0.2380 \\
& Qwen2.5-VL-7B~\cite{Qwen2.5-VL}      & 0.4164 & 0.5420     & 0.3214 & 0.4027 & 0.4619   & 0.2889 & 0.2731 & 0.3866\\
& Qwen2.5-VL-32B~\cite{Qwen2.5-VL}     & 0.4588 & 0.5564     & 0.2814 & 0.4200 & 0.5003   & 0.3170 & 0.2917 & 0.4036 \\
& InternVL-14B~\cite{internvl2024}            & 0.4288     & 0.5583     & 0.4311  & 0.4349    & 0.5139     & 0.3625     & 0.3490        & 0.4398 \\
\rowcolor{gray!20}
& EyePCR-MLLM (7B)(Ours)       &  \textcolor{red}{\textbf{0.8991}}    & \textcolor{red}{0.8768}    &  \textcolor{red}{0.6872}     & \textcolor{red}{0.6776}    & \textcolor{red}{0.8339}  & \textcolor{red}{0.5452}   & \textcolor{red}{0.6683}    & \textcolor{red}{0.7412} \\
\midrule
\multirow{2}{*}{\rotatebox[origin=c]{90}{\small \textbf{Expert}}} 
& Ophthalmology Trainee            & 0.6240     & 0.6500     & 0.9165     & 0.6275     & 0.8125     & 0.6000     & 0.3667     & 0.6567 \\
& Ophthalmic Surgeon   & 0.7917     & \textbf{0.9445}     & \textbf{0.9792}     & \textbf{0.8959}     & \textbf{0.9688}     & \textbf{0.8667}     & \textbf{0.9167}     & \textbf{0.9090} \\
\bottomrule
\end{tabular}
}
\label{tab:MCQ}
\end{table*}

\noindent\textbf{EyePCR Dataset Is Challenging.} \quad 
EyePCR poses a substantial challenge to MLLMs while remaining clinically valid. This is evidenced by the large performance gap between human experts and MLLMs. Ophthalmic surgeons achieve 0.9090 accuracy on average, confirming the answerability and clarity of the questions. But leading commercial models like GPT-4.1 and Gemini 2.5 Pro reach only 0.7069 and \textcolor{blue}{0.7138}, respectively, and most open-source models perform below 0.45. This discrepancy shows that the EyePCR is not ambiguous or ill-posed, but rather demands precise visual parsing and domain-specific understanding ability not yet well-captured by general-purpose MLLMs. 


\noindent\textbf{EyePCR Largely Enhances MLLMs.} \quad 
Fine-tuning on the EyePCR dataset significantly improves the domain-specific perceptual capabilities of vision-language models. Compared to its base model Qwen2.5-VL-7B, \textbf{EyePCR-MLLM} achieves a substantial gain on average MCQ accuracy, improving from 0.3866 to \textcolor{red}{0.7412}, demonstrating the effectiveness of the EyePCR framework and associated dataset. It consistently outperforms open-source baselines, including larger models such as Qwen2.5-VL-32B and InternVL-14B. Notably, it also surpasses leading commercial MLLMs such as GPT-4.1 (0.7069) and Gemini 2.5 Pro (\textcolor{blue}{0.7138}), despite having a significantly smaller model (7B vs. 32B+). When compared to human participants, EyePCR-MLLM exceeds \textbf{Ophthalmology Trainee} (0.6567) across all perceptual views, and narrows the gap with the \textbf{Ophthalmic Surgeon} (0.9090), particularly in visually grounded categories such as \textbf{V1} and \textbf{V5}, highlighting the value of the EyePCR dataset in bridging the clinical gap between generic MLLMs and expert-level surgical perception. 



\noindent\textbf{MLLMs Still Lag behind Experts.} \quad 
As shown in Tab.~\ref{tab:perception-mcq-accuracy}, while \textbf{EyePCR-MLLM} and \textbf{Gemini 2.5 Pro} outperform \textbf{Ophthalmology Trainee}, they lag behind \textbf{Ophthalmic Surgeon}. It shows that though capable of recognizing common anatomical and procedural cues, current models lack the precision and consistency to handle subtle visual signals, or ambiguous contexts. The human-AI performance gap shows that clinically reliable perception requires not only large-scale data and model size, but also deeper integration of surgical semantics and domain-specific visual reasoning.

\begin{table*}[!t]
\centering
\large
\renewcommand{\arraystretch}{1.5}
\caption{\textbf{OEQs} performance in BLEU-4 (B-4) and ROUGE-L (R-L) across \textbf{Perception (P)}, \textbf{Comprehension (C)}, and \textbf{Reasoning (D1–D4)}, alongside average and ChatGPT-based rubric scores.  
\textcolor{blue}{\textbf{Blue}}: highest among \textbf{commercial models}; 
\textcolor{red}{\textbf{Red}}: highest among \textbf{open-source models}; 
\textbf{Bold}: best overall. 
}
\vspace{0.3em}
\label{tab:vqa-metric-final}
\resizebox{\textwidth}{!}{
\begin{tabular}{cl|cc|cc|cccccccc|cc|c}
\toprule
& \multirow{3}{*}{\textbf{Participants}} 
& \multicolumn{2}{c|}{\textbf{P}} 
& \multicolumn{2}{c|}{\textbf{C}} 
& \multicolumn{8}{c|}{\textbf{R}} 
& \multicolumn{2}{c|}{\textbf{Avg}} 
& \multirow{2}{*}{\textbf{ChatGPT}} \\
\cmidrule(lr){3-4} \cmidrule(lr){5-6} \cmidrule(lr){7-14} \cmidrule(lr){15-16}
& & \textbf{B-4} & \textbf{R-L} 
& \textbf{B-4} & \textbf{R-L} 
& \textbf{D1 B-4} & \textbf{D1 R-L} 
& \textbf{D2 B-4} & \textbf{D2 R-L} 
& \textbf{D3 B-4} & \textbf{D3 R-L} 
& \textbf{D4 B-4} & \textbf{D4 R-L} 
& \textbf{B-4} & \textbf{R-L} & \textbf{(0$\rightarrow$5)} \\
\midrule
\multirow{6}{*}{\rotatebox[origin=c]{90}{\textbf{Commercial}}} 
& Gemini 2.0 Flash~\cite{google2024gemini}     & 0.0624 & 0.3783 & 0.0603 & 0.2926 & 0.0329 & 0.1956 & 0.0513 & 0.2499 & 0.0338 & 0.2105 & 0.0392 & 0.2214 & 0.0467 & 0.2581 & 3.2417\\
& Claude 3.7~\cite{claude2024sonnet}     & 0.0497 & \textcolor{blue}{\textbf{0.4207}} & 0.0518 & 0.2888 & 0.0375 & 0.2371 & 0.0652 & 0.2850 & 0.0506 & 0.2818 & 0.0499 & 0.2736 & 0.0508 & 0.2978 & 3.5261\\
& Gemini 2.5 Flash~\cite{google2024gemini} & 0.0568 & 0.3993 & 0.0593 & 0.2871 & 0.0448 & 0.2567 & 0.0649 & 0.2790 & 0.0576 & 0.2772 & 0.0482 & 0.2615 & 0.0553 & 0.2935 & 3.5048\\
& GPT-o3~\cite{openai2024gpt3.5}          & 0.0449 & 0.3757 & 0.0475 & 0.2820 & 0.0469 & 0.2722 & 0.0470 & 0.2606 & 0.0484 & 0.2798 & 0.0446 & 0.2686 & 0.0466 & 0.2898 & \textcolor{blue}{3.9349}\\
& GPT-4.1~\cite{openai2024gpt41}           & 0.0398 & 0.3069 & 0.0513 & 0.2696 & 0.0365 & 0.2297 & 0.0507 & 0.2636 & 0.0445 & 0.2451 & 0.0411 & 0.2351 & 0.0440 & 0.2583 & 3.8405\\
& Gemini 2.5 Pro~\cite{google2024gemini}  & \textcolor{blue}{0.0685} & 0.4060 & \textcolor{blue}{\textbf{0.0697}} & \textcolor{blue}{\textbf{0.3200}} & \textcolor{blue}{\textbf{0.0573}} & \textcolor{blue}{\textbf{0.2877}} & \textcolor{blue}{\textbf{0.0893}} & \textcolor{blue}{\textbf{0.3213}} & \textcolor{blue}{\textbf{0.0829}} & \textcolor{blue}{\textbf{0.3245}} & \textcolor{blue}{\textbf{0.0709}} & \textcolor{blue}{\textbf{0.3163}} & \textcolor{blue}{\textbf{0.0731}} & \textcolor{blue}{\textbf{0.3293}} & 3.6949\\
\midrule
\multirow{6}{*}{\rotatebox[origin=c]{90}{\textbf{Open-Source}}} 
& VideoLLaMA3-7B~\cite{VideoLLaMA3-7B} & 0.0174     & 0.1351     & 0.0211     & 0.1546     & 0.0189     & 0.1398     & 0.0174     & 0.1469     & 0.0198     & 0.1364     & 0.0157     & 0.1235     & 0.0183     & 0.1393  & 2.3605\\
& LLaVA-Video-7B~\cite{LLaVA-Video-7B}    & 0.0060 & 0.0598 & 0.0087 & 0.0808 & 0.0077 & 0.0683 & 0.0091 & 0.0965 & 0.0080 & 0.0754 & 0.0081 & 0.0766 & 0.0079 & 0.0762 & 1.8765\\
& Qwen2.5-VL-7B~\cite{Qwen2.5-VL}    & 0.0049 & 0.0465 & 0.0062 & 0.0597 & 0.0076 & 0.0653 & 0.0072 & 0.0685 & 0.0060 & 0.0563 & 0.0056 & 0.0527 & 0.0063 & 0.0582 & 2.1659\\
& Qwen2.5-VL-32B~\cite{Qwen2.5-VL}   & 0.0416 & 0.2995 & 0.0495 & 0.2631 & 0.0371 & 0.2124 & 0.0398 & 0.2278 & 0.0315 & 0.1938 & 0.0288 & 0.1902 & 0.0381 & 0.2311 & 2.7163\\
& InternVL-14B~\cite{internvl2024}           & 0.0127     & 0.1182     & 0.0158     & 0.1287     & 0.0143     & 0.1155     & 0.0171     & 0.1408     & 0.0128     & 0.1128     & 0.0137     & 0.1120     & 0.0144     & 0.1213 & 2.5446\\
\rowcolor{gray!20}
& EyePCR-MLLM (7B)(\small Ours) & \textcolor{red}{\textbf{0.1099}} & \textcolor{red}{0.3795} & \textcolor{red}{0.0513} & \textcolor{red}{0.2513} & \textcolor{red}{0.0510} & \textcolor{red}{0.2355} & \textcolor{red}{0.0703} & \textcolor{red}{0.2792} & \textcolor{red}{0.0645} & \textcolor{red}{0.2638} & \textcolor{red}{0.0597} & \textcolor{red}{0.2612} & \textcolor{red}{0.0678} & \textcolor{red}{0.2784} & \textcolor{red}{3.0062}\\
\midrule
\multirow{2}{*}{\rotatebox[origin=c]{90}{\textbf{Expert}}} 
& Ophthalmology Trainee      & 0.0092     & 0.0808     & 0.0057     & 0.0927     & 0.0010     & 0.0105     & 0.0075     & 0.0782     & 0.0012     & 0.1253     & 0.0142     & 0.1581     & 0.0065     & 0.0909 & 3.1697\\
& Ophthalmic Surgeon     & 0.0108     & 0.1431     & 0.0085     & 0.1245     & 0.0054     & 0.0633     & 0.0080     & 0.1028     & 0.0141     & 0.1617     & 0.0230     & 0.2769     & 0.0116     & 0.1454 & \textbf{4.1583} \\
\bottomrule
\end{tabular}
}
\end{table*}

\subsection{Evaluation on OEQs for Perception, Comprehension and Reasoning} 
We further evaluate models on Open-Ended Questions (OEQs) across the three cognitive stages: Perception (P), Comprehension (C), and Reasoning (D1–D4). As shown in Tab.~\ref{tab:vqa-metric-final}, we report BLEU-4, ROUGE-L scores, and GPT-4.1–based rubric evaluations. 
Several key findings emerge: 

\noindent\textbf{OEQs in EyePCR Are Challenging Yet Valid.} \quad
Compared to MCQs, all models exhibit significantly lower scores on OEQs, \eg, Gemini 2.5 Pro achieves the highest BLEU-4 of 0.0731, while most open-source models fall below 0.04, highlighting the increased complexity of generative surgical understanding. Interestingly, human experts also score poorly under lexical metrics, \eg, BLEU-4 of Ophthalmic Surgeon is 0.0116, not due to a lack of domain knowledge, but because expert responses are often concise and semantically precise, lacking n-gram overlap with references. To account for this, we incorporate rubric-based evaluation that scores answers based on factuality, specificity, and completeness. Under this rubric, the Ophthalmic Surgeon achieves the highest score of 4.16, confirming the clinical validity and answerability of EyePCR's OEQs. 

\noindent\textbf{EyePCR Enhances MLLMs.} \quad
We utilize the EyePCR framework to fine-tune Qwen2.5-VL-7B to obtain \textbf{EyePCR-MLLM}. Despite having the same model size, EyePCR-MLLM shows substantial improvements, \eg, BLEU-4 rises from 0.0063 to 0.0678 and ROUGE-L from 0.0582 to 0.2784. These gains are consistent across the three stages and four reasoning subtasks, demonstrating that EyePCR’s structured multi-view supervision and clinically grounded QA generation largely enhance model alignment with surgical cognition. EyePCR-MLLM surpasses larger open-source models like Qwen2.5-VL-32B and achieves performance comparable to commercial systems like ChatGPT4.1. 



\noindent\textbf{Rubric Scores Better Assess Semantic Understanding.} \quad 
The disparity between lexical metrics and rubric evaluations underscores the limitations of BLEU and ROUGE in clinical QA. These metrics rely on n-gram overlap and tend to reward surface-level lexical similarity~\cite{singhal2025toward, alonso2024medexpqa}, which penalizes semantically concise responses, particularly those written by experts. In contrast, rubric scores assess model outputs based on high-level criteria such as factuality, specificity, and completeness, allowing for valid answers that differ lexically from the reference but are clinically accurate. As a result, while Ophthalmic Surgeon achieves low BLEU-4, they attain the highest rubric score, surpassing all models, demonstrating that rubric evaluation better aligns with domain expertise and is more meaningful for assessing free-text responses in safety-critical medical settings. 
Integrating semantic-aware evaluation is thus crucial for measuring generative quality in high-stakes medical applications.

\section{Conclusion and Future Work} 
\label{sec:Conclusion} 
We present \textbf{EyePCR}, the first comprehensive benchmark for evaluating surgical cognitive intelligence in ophthalmic surgery, which integrates multi-view fine-grained attributes, structured knowledge graphs, and clinically grounded reasoning tasks, forming a unified \textbf{Perception}–\textbf{Comprehension}–\textbf{Reasoning} evaluation paradigm. 
More than 210k VQAs are generated to offer a complete suite 
for evaluating a models' cognitive abilities in  \textbf{Perception}, \textbf{Comprehension}, and \textbf{Reasoning}, enabling precise perception on fine-grained attributes, aligning with procedural semantics, and simulating expert-level clinical inference. Extensive experiments across 12 MLLMs reveal huge performance disparities: while commercial models demonstrate strong perceptual fluency, their reasoning abilities remain shallow without domain adaptation. 
Our domain-specialized EyePCR-MLLM sets new benchmarks among open models and approaches the performance of leading commercial solutions, affirming the value of structured surgical representation and task-specific fine-tuning. 

Looking forward, EyePCR not only provides a robust benchmark for assessing surgical AI models, but also lays a foundation for future development of clinically reliable, interpretable, and decision-supportive multimodal foundation models. 
Besides our own research efforts in MLLMs, we hope this work could inspire broader efforts in medical multimodal learning, especially toward the development of open, transparent, and generalizable surgical video understanding systems.

\bibliographystyle{unsrt}
\bibliography{references}

\clearpage
\newpage

\end{document}